\documentclass[10pt,twocolumn,letterpaper]{article}

\usepackage{cvpr}
\usepackage{times}
\usepackage{epsfig}
\usepackage{graphicx}
\usepackage{amsmath}
\usepackage{amssymb}
\usepackage{array}
\usepackage{tabulary}
\newcolumntype{K}[1]{>{\centering\arraybackslash}m{#1}}
\usepackage{subcaption}
\usepackage{cite}
\usepackage{url}

\usepackage[breaklinks=true,bookmarks=false, backref=page]{hyperref}

\cvprfinalcopy 

\begin{document}

\title{SWA Object Detection}
\author{Haoyang Zhang$^{1}$, Ying Wang$^{2}$, Feras Dayoub$^{1}$, Niko S\"underhauf$^{1}$\thanks{The authors acknowledge the continued support from Queensland University of Technology (QUT) through the Centre for Robotics.} \\
$^{1}$Australian Centre for Robotic Vision, QUT Centre for Robotics, Queensland University of Technology \\
$^{2}$University of Queensland\\
{\tt\small \{h202.zhang, feras.dayoub, niko.suenderhauf\}@qut.edu.au, ying.wang@uq.edu.au}
}
\renewcommand\footnotemark{}

\maketitle

\begin{abstract}
Do you want to improve 1.0 AP for your object detector without any inference cost and any change to your detector? Let us tell you such a recipe. It is surprisingly simple: \textbf{train your detector for an extra 12 epochs using cyclical learning rates and then average these 12 checkpoints as your final detection model}. This potent recipe is inspired by Stochastic Weights Averaging (\textbf{SWA}), which is proposed in~\cite{SWA} for improving generalization in deep neural networks. We found it also very effective in object detection. In this technique report, we systematically investigate the effects of applying SWA to object detection as well as instance segmentation. Through extensive experiments, we discover the aforementioned workable policy of performing SWA in object detection, and we consistently achieve $\sim$1.0 AP improvement over various popular detectors on the challenging COCO benchmark, including Mask RCNN, Faster RCNN, RetinaNet, FCOS, YOLOv3 and VFNet. We hope this work will make more researchers in object detection know this technique and help them train better object detectors. Code is available at: https://github.com/hyz-xmaster/swa\_object\_detection .

\end{abstract}

\section{Introduction}
Thanks to the big success in deep learning, object detection has made great progress in recent years. In 2015, Faster RCNN~\cite{fasterRCNN} only achieved 21.9 AP on COCO test-dev~\cite{COCO}, whereas this number has improved to about 61.0 in 2020 on the latest COCO leaderboard~\cite{COCO_leaderboard}. Nonetheless, we can see that the evolution of object detection is becoming slow because the feature representation learning capacity of deep networks has almost been squeezed dry. According to the report of 2020 COCO+LVIS Joint Recognition Challenge~\cite{COCOLVIS}, the performance of object detection (instance segmentation track) on COCO has reached saturation these two years, indicating that it is becoming harder to improve object detection performance. Even though researchers rack their brains, trying to design better detector modules, they may find it difficult to improve the performance further by 1.0 AP on the challenging COCO benchmark in future.  

On the other hand, we have recently found a very simple but effective way of enhancing object detectors in our research, which we are excited to share with the community. \textbf{You only need to train your detector for an extra 12 epochs using cyclical learning rates and then average these 12 checkpoints as your final detection model.} As a result, you can get $\sim$1.0 AP improvement on the challenging COCO benchmark. Since this technique only incurs some training overhead, you do not need to be worried about any inference cost and any change to your detectors.

This technique is developed in~\cite{SWA} for improving generalization in deep networks and is termed as Stochastic Weights Averaging (\textbf{SWA}). We attempted it in our research of object detection and was surprised by its effectiveness in improving our object detector, VarifocalNet~\cite{VFNet} or VFNet for short. We found that rare work~\cite{koohbanani2019nuclear} of object detection had adopted this technique. Therefore, we did a systematic study of the effects of applying SWA to object detection. We first selected Mask RCNN~\cite{maskRCNN} as our study object detector due to its representativeness and popularity. We then tried different training strategies and discovered the aforementioned workable policy of performing SWA in object detection. With this policy, through extensive experiments, we found SWA could improve $\sim$1.0 AP on the COCO benchmark for various object detectors, including Mask RCNN~\cite{maskRCNN}, Faster RCNN~\cite{fasterRCNN}, RetinaNet~\cite{retinaNet}, FCOS~\cite{FCOS}, YOLOv3\cite{YOLOv3} and our VFNet~\cite{VFNet}. This makes us excited to share this discovery and hope this work will be helpful to the community in training better object detectors.

\section{SWA}
We briefly describe what SWA is and why it works. For more details, please refer to the SWA paper~\cite{SWA}, its blog~\cite{SWA_blog} or the related tutorial~\cite{SWA_tutorial}.

Simply put, SWA is the averaging of multiple checkpoints along the optimization trajectory of SGD with a high constant learning rate or cyclical learning rates. Let $w_{i}$ denote the checkpoint of epoch $i$. In conventional SGD, the checkpoint of the last epoch $w_{n}$ or the best one on the validation $w_{i}^{*}$ is generally selected as the final model. By contrast, in SWA, the average of multiple checkpoints $\bar w = 1/(n-m+1) \sum_{i=m}^{n} w_{i}$ is adopted as the final model. 

\begin{figure}[t]
	\begin{center}
		\includegraphics[width=0.95\columnwidth]{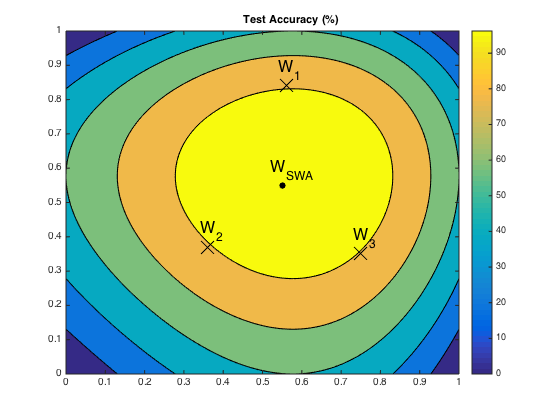}
	\end{center}
	\vspace{-8mm}
	\caption{Illustration of SWA. $W_{1}$, $W_{2}$ and $W_{3}$ represent different SGD solutions. $W_{SWA}$ indicates the SWA solution that is the average of those SGD solutions.}
	\label{fig:swa}
\vspace{-2mm}
\end{figure}

Why does this simple method work? The authors argue that SGD usually converges to a solution that is on the periphery of the space of a set of good weights (like $W_{1}$ in Figure~\ref{fig:swa}) and this solution normally generalizes worse than those that are centered in the space. Running SGD with a cyclical or a high constant learning rate schedule allows SGD optimization to explore multiple points near the boundary of the flat weights space corresponding to deep neural networks with high accuracy, shown as $W_{1}$, $W_{2}$ and $W_{3}$ in Figure~\ref{fig:swa}. Then, by averaging these points, SWA can find a more centred solution $W_{SWA}$ that has substantially better generalization.

In practice, there are two main questions to answer for applying SWA in training an object detector. First, what learning rate schedule should we use for SWA training from epoch $m$ to epoch $n$? Use a high constant learning rate or cyclical learning rates? Second, how many checkpoints should we average? That is, how many epochs should we train for SWA? In this report, we answer these questions through extensive experiments.

\section{Experiments}

In this section, we conduct a series of experiments to investigate the effects and discover an appropriate way of applying SWA to object detection. 

\vspace{-4mm}
\paragraph{Dataset and Evaluation Metrics.}
We do the experiments on the widely-used MS COCO 2017 dataset~\cite{COCO}. We train detectors on the \texttt{train2017} split and report results on the \texttt{val2017} split. We adopt the standard COCO-style Average Precision (AP) as the evaluation metrics. 

\vspace{-4mm}
\paragraph{Implementation and Training Details.} 
We rely on MMDetection~\cite{mmdetection} for our experiments. We use 8 V100 GPUs for training with a total batch size of 16 (2 images per GPU). For convenience, we describe 1x and 2x training schedules~\cite{Detectron2} here. 1x schedule means a model is trained for 12 epochs and the initial learning rate decreases by a factor of 10 at epoch 9 and epoch 12 respectively, and 2x schedule means a model is trained for 24 epochs and the initial learning rate decreases by a factor of 10 at epoch 17 and epoch 23 respectively. For brevity, we also describe here the naming rule of object detectors used in this report. Take MaskRCNN-R101-2x-0.02-0.0002-40.8-36.6 as an example for explanation. It means that the \textbf{pre-trained} detector, Mask RCNN, has a ResNet-101~\cite{ResNet} backbone, is trained under 2x schedule with the initial learning rate 0.02 and ending learning rate 0.0002, and achieves 40.8 bbox AP and 36.6 mask AP on COCO val2017 respectively.

\begin{figure}[t]
	\begin{center}
		\includegraphics[width=0.95\columnwidth]{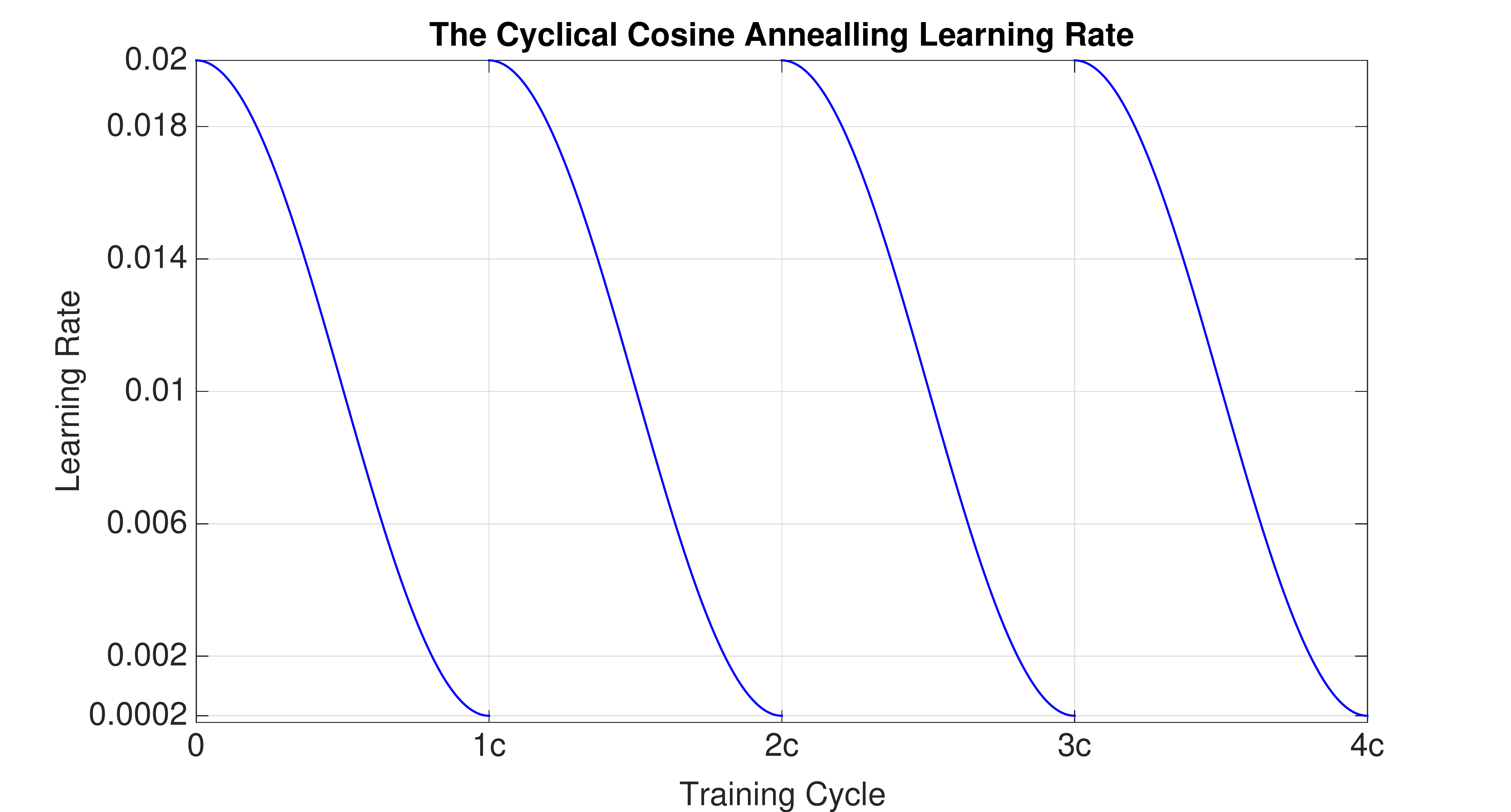}
	\end{center}
	\vspace{-3mm}
	\caption{Illustration of cyclical cosine annealing learning rates. In each cycle, the learning rate decreases at each iteration from the initial learning rate $lr_{max}$ (0.02 in this case) to the ending learning rate $lr_{min}$ (0.0002 in this case).}
	\label{fig:cyclical_lr}
\vspace{-2mm}
\end{figure}

\begin{table*}[tpb]
    \begin{center}
        \begin{tabular}{ K{4.0em} | K{1.5em} K{1.5em} K{1.5em} K{1.5em} K{1.5em} K{1.5em} K{1.5em} K{1.5em} K{1.5em} K{1.5em} K{1.5em} K{1.5em} | K{1.8em} K{2.0em} K{2.0em} K{2.0em}} 
            \hline
               & 1 & 2 & 3 &  4 & 5 & 6 & 7 &  8 & 9 & 10 & 11 & 12 & SWA 1-6 & SWA 1-12 & SWA 1-24 & SWA 1-48 \\
            \hline
             detector & \multicolumn{12}{c|}{\textbf{MaskRCNN-R101-2x-0.02-0.0002-40.8-36.6}} \\
            \hline
             strategy & \multicolumn{12}{c|}{fixed lr = 0.02, 24 epochs} \\
            \hline
             bbox AP & 32.8 & 33.3 & 34.1 & 33.9 & 33.6 & 33.9 & 34.4 & 34.1 & 34.4 & 34.0 & 34.4 & 33.7 &  &  &  \\
             mask AP & 30.2 & 30.9 & 31.6 & 31.1 & 30.9 & 31.8 & 31.9 & 31.7 & 31.8 & 31.6 & 31.4 & 31.1 &  &  &  \\
             \hline
             bbox AP & 34.0 & 34.0 & 34.1 & 33.6 & 34.8 & 34.3 & 34.3 & 34.5 & 34.4 & 34.7 & 34.5 & 34.0 & 39.5 & 40.3 & 40.6 & - \\
             mask AP & 31.4 & 31.6 & 31.3 & 30.9 & 32.3 & 31.7 & 31.6 & 31.9 & 32.1 & 32.1 & 32.1 & 31.6 & 35.9 & 36.5 & 36.8 & - \\
             
            \hline
             strategy & \multicolumn{12}{c|}{fixed lr = 0.002, 24 epochs} \\
            \hline
             bbox AP & 40.0 & 40.1 & 40.1 & 39.6 & 39.5 & 39.4 & 39.1 & 38.9 & 38.9 & 38.4 & 38.1 & 38.3 &  &  &  \\
             mask AP & 36.0 & 36.1 & 36.1 & 35.6 & 35.6 & 35.5 & 35.4 & 35.1 & 35.1 & 34.8 & 34.4 & 34.7 &  &  &  \\
             \hline
             bbox AP & 38.5 & 37.9 & 37.9 & 37.6 & 37.6 & 37.2 & 37.2 & 37.6 & 37.2 & 36.9 & 36.7 & 36.7 & 40.6 & 40.5 & 40.0 & - \\
             mask AP & 34.8 & 34.2 & 34.3 & 34.2 & 34.2 & 33.7 & 33.6 & 34.0 & 33.7 & 33.5 & 33.4 & 33.4 & 36.5 & 36.4 & 36.0 & - \\
             
            \hline
             strategy & \multicolumn{12}{c|}{fixed lr = 0.0002, 24 epochs} \\
            \hline
             bbox AP & 40.7 & 40.6 & 40.6 & 40.6 & 40.5 & 40.6 & 40.4 & 40.4 & 40.3 & 40.3 & 40.4 & 40.3 &  &  &  \\
             mask AP & 36.6 & 36.6 & 36.5 & 36.5 & 36.4 & 36.5 & 36.4 & 36.4 & 36.3 & 36.2 & 36.3 & 36.2 &  &  &  \\
             \hline
             bbox AP & 40.3 & 40.1 & 40.1 & 40.1 & 40.1 & 40.0 & 40.0 & 39.9 & 39.9 & 39.9 & 39.8 & 39.7 & 40.7 & 40.5 & 40.3 & - \\
             mask AP & 36.2 & 36.1 & 36.1 & 36.0 & 36.0 & 35.9 & 35.9 & 35.8 & 35.9 & 35.8 & 35.8 & 35.7 & 36.6 & 36.4 & 36.2 & - \\
             
             \hline
             strategy & \multicolumn{12}{c|}{cyclical lr = (0.01, 0.0001), cycle length = 1 epoch, 48 epochs} \\
             \hline
             bbox AP & 40.8 & 40.8 & 40.8 & 40.8 & 40.7 & 40.7 & 40.7 & 40.9 & 40.7 & 40.5 & 40.5 & 40.7 &  &  &  \\
             mask AP & 36.7 & 36.7 & 36.6 & 36.7 & 36.5 & 36.7 & 36.7 & 36.8 & 36.5 & 36.3 & 36.4 & 36.5 &  &  &  \\
             \hline
             bbox AP & 40.6 & 40.3 & 40.4 & 40.4 & 40.2 & 40.2 & 40.5 & 40.3 & 40.2 & 40.1 & 40.2 & 39.9 & & & & \\
             mask AP & 36.5 & 36.3 & 36.2 & 36.4 & 36.1 & 36.1 & 36.3 & 36.2 & 36.1 & 35.9 & 36.2 & 35.9 & & & & \\
             \hline
             bbox AP & 39.9 & 39.7 & 39.8 & 39.7 & 40.0 & 39.7 & 39.7 & 39.9 & 39.8 & 39.7 & 39.6 & 39.4 &  &  &  \\
             mask AP & 36.0 & 35.8 & 35.9 & 35.8 & 36.0 & 35.8 & 35.7 & 35.9 & 35.8 & 35.7 & 35.6 & 35.5 &  &  &  \\
             \hline
             bbox AP & 39.5 & 39.5 & 39.4 & 39.3 & 39.5 & 39.3 & 39.1 & 39.2 & 39.2 & 39.0 & 39.4 & 39.0 & 41.3 & 41.5 & 41.6 & 41.3\\
             mask AP & 35.6 & 35.6 & 35.4 & 35.4 & 35.6 & 35.5 & 35.2 & 35.4 & 35.3 & 35.1 & 35.4 & 35.1 & 37.2 & 37.3 & 37.3 & 37.1\\
             
             \hline
             strategy & \multicolumn{12}{c|}{cyclical lr = (0.02, 0.0002), cycle length = 1 epoch, 48 epochs} \\
             \hline
             bbox AP & 40.7 & 40.4 & 40.3 & 40.7 & 40.6 & 40.7 & 40.6 & 40.6 & 40.7 & 40.6 & 40.7 & 40.6 &  &  &  \\
             mask AP & 36.6 & 36.4 & 36.4 & 36.7 & 36.6 & 36.7 & 36.6 & 36.6 & 36.6 & 36.7 & 36.7 & 36.6 &  &  &  \\
             \hline
             bbox AP & 40.9 & 40.6 & 40.6 & 40.8 & 40.7 & 40.5 & 40.7 & 40.8 & 40.9 & 40.7 & 40.8 & 40.7 & & &  \\
             mask AP & 36.7 & 36.6 & 36.5 & 36.6 & 36.5 & 36.6 & 36.6 & 36.6 & 36.7 & 36.6 & 36.6 & 36.7 & &  &  \\
            \hline
             bbox AP & 40.6 & 40.6 & 40.6 & 40.7 & 40.6 & 40.5 & 40.7 & 40.7 & 40.7 & 40.7 & 40.7 & 40.6 &  &  &  \\
             mask AP & 36.8 & 36.7 & 36.6 & 36.7 & 36.5 & 36.6 & 36.6 & 36.6 & 36.6 & 36.6 & 36.6 & 36.6 &  &  &  \\
             \hline
             bbox AP & 40.6 & 40.5 & 40.7 & 40.5 & 40.6 & 40.7 & 40.6 & 40.7 & 40.7 & 40.7 & 40.6 & 40.6 & 41.5 & \textbf{41.7} & \textbf{41.7} & \textbf{41.7} \\
             mask AP & 36.6 & 36.6 & 36.6 & 36.5 & 36.5 & 36.5 & 36.6 & 36.5 & 36.8 & 36.6 & 36.5 & 36.6 & 37.3 & 37.4 & 37.5 & \textbf{37.6} \\
             
             \hline
             detector & \multicolumn{12}{c|}{\textbf{MaskRCNN-R101-16e-0.02-0.02-33.4-30.8}} \\
            \hline
             strategy & \multicolumn{12}{c|}{cyclical lr = (0.02, 0.0002), cycle length = 1 epoch, 12 epochs} \\
             \hline
             bbox AP & 40.7 & 40.5 & 40.7 & 40.9 & 40.7 & 40.5 & 40.8 & 40.8 & 40.7 & 40.8 & 40.6 & 40.8 & 41.5 & \textbf{41.7} & - & - \\
             mask AP & 36.7 & 36.5 & 36.7 & 36.9 & 36.7 & 36.4 & 36.7 & 36.6 & 36.5 & 36.7 & 36.5 & 36.7 & 37.3 & \textbf{37.4} & - & - \\
            \hline
        \end{tabular}
    \end{center}
    \vspace{-3mm}
\caption{Performances of each SGD epoch of the further trained Mask RCNN and corresponding SWA models on the COCO val2017.  Strategy means the learning rate schedule we adopted for training the pre-trained Mask RCNN model (see text). SWA e1-e2 means the model obtained by averaging the checkpoints from epoch e1 to epoch e2.  }
\label{table:maskrcnn}
\vspace{-4mm}
\end{table*}

\begin{table*}[tpb]
    \begin{center}
        \begin{tabular}{ K{4.0em} | K{1.2em} K{1.2em} K{1.2em} K{1.2em} K{1.2em} K{1.2em} K{1.2em} K{1.2em} K{1.2em} K{1.2em} K{1.2em} K{1.2em} | K{3.8em} K{4.2em}}
            \hline
               & 1 & 2 & 3 &  4 & 5 & 6 & 7 &  8 & 9 & 10 & 11 & 12 & SWA 1-6 & SWA 1-12 \\
            \hline
             detector & \multicolumn{12}{c|}{\textbf{MaskRCNN-R50-1x-0.02-0.0002-38.2-34.7}} \\
             \hline
             strategy & \multicolumn{12}{c|}{cyclical lr = (0.02, 0.0002), cycle length = 1 epoch, 12 epochs} \\
             \hline
             bbox AP & 37.8 & 38.0 & 38.0 & 38.1 & 38.4 & 38.1 & 38.3 & 38.3 & 38.5 & 38.3 & 38.4 & 38.6 & 38.8 +0.6 & \textbf{39.1} +0.9 \\
             mask AP & 34.4 & 34.6 & 34.5 & 34.7 & 34.8 & 34.7 & 34.7 & 34.8 & 34.9 & 34.9 & 34.9 & 35.0 & 35.2 +0.5 & \textbf{35.5} +0.8 \\
             
            \hline
             detector & \multicolumn{12}{c|}{\textbf{MaskRCNN-R101-1x-0.02-0.0002-40.0-36.1}} \\
             \hline
             strategy & \multicolumn{12}{c|}{cyclical lr = (0.02, 0.0002), cycle length = 1 epoch, 12 epochs} \\
             \hline
             bbox AP & 39.8 & 39.7 & 39.8 & 40.0 & 40.0 & 40.1 & 40.1 & 40.4 & 40.2 & 40.2 & 40.4 & 40.4 & 40.7 +0.7 & \textbf{41.0} +1.0 \\
             mask AP & 36.0 & 35.8 & 36.1 & 36.2 & 36.2 & 36.2 & 36.2 & 36.4 & 36.2 & 36.3 & 36.3 & 36.4 & 36.8 +0.7 & \textbf{37.0} +0.9 \\

            \hline
        \end{tabular}
    \end{center}
    \vspace{-3mm}
\caption{Performances of SWA Mask RCNN.}
\label{table:maskrcnn_res}
\end{table*}

\begin{table*}[tpb]
    \begin{center}
        \begin{tabular}{ K{4.0em} | K{1.0em} K{1.0em} K{1.0em} K{1.0em} K{1.0em} K{1.0em} K{1.0em} K{1.0em} K{1.0em} K{1.0em} K{1.0em} K{1.2em} | K{3.8em} K{4.2em} K{4.2em}}
            \hline
             & 1 & 2 & 3 &  4 & 5 & 6 & 7 &  8 & 9 & 10 & 11 & 12 & SWA 1-6 & SWA 1-12 & SWA 1-24 \\
            \hline
             detector & \multicolumn{12}{c|}{\textbf{FasterRCNN-R50-1x-0.02-0.0002-37.4}} \\
             \hline
             strategy & \multicolumn{12}{c|}{cyclical lr = (0.02, 0.0002), cycle length = 1 epoch, 12 epochs} \\
             \hline
             bbox AP & 37.0 & 37.1 & 37.2 & 37.4 & 37.5 & 37.6 & 37.6 & 37.6 & 37.8 & 37.9 & 37.8 & 37.9 & 37.9 +0.5 & \textbf{38.4} +1.0 & - \\
             
            \hline
             detector & \multicolumn{12}{c|}{\textbf{FasterRCNN-R101-1x-0.02-0.0002-39.4}} \\
             \hline
             strategy & \multicolumn{12}{c|}{cyclical lr = (0.02, 0.0002), cycle length = 1 epoch, 12 epochs} \\
             \hline
             bbox AP & 39.1 & 39.0 & 39.3 & 39.3 & 39.4 & 39.5 & 39.4 & 39.5 & 39.6 & 39.6 & 39.5 & 39.7 & 39.9 +0.5 & \textbf{40.3} +0.9 & - \\
             
             \hline
             detector & \multicolumn{12}{c|}{\textbf{FasterRCNN-R101-2x-0.02-0.0002-39.8}} \\
             \hline
             strategy & \multicolumn{12}{c|}{cyclical lr = (0.02, 0.0002), cycle length = 1 epoch, 24 epochs} \\
             \hline
             bbox AP & 39.8 & 39.7 & 39.7 & 39.9 & 39.7 & 39.8 & 39.8 & 39.8 & 39.9 & 39.7 & 39.7 & 39.7 &  &  & \\
             bbox AP & 39.9 & 39.7 & 39.9 & 39.9 & 39.8 & 39.8 & 39.8 & 39.9 & 39.8 & 39.7 & 39.8 & 40.0 & 40.6 +0.8 & 40.7 +0.9 & \textbf{40.9} +1.1\\

            \hline
        \end{tabular}
    \end{center}
    \vspace{-3mm}
\caption{Performances of SWA Faster RCNN.}
\label{table:fasterrcnn}
\end{table*}

\begin{table*}[tpb]
    \begin{center}
        \begin{tabular}{ K{4.0em} | K{1.0em} K{1.0em} K{1.0em} K{1.0em} K{1.0em} K{1.0em} K{1.0em} K{1.0em} K{1.0em} K{1.0em} K{1.0em} K{1.2em} | K{3.8em} K{4.2em} K{4.2em}}
            \hline
             & 1 & 2 & 3 &  4 & 5 & 6 & 7 &  8 & 9 & 10 & 11 & 12 & SWA 1-6 & SWA 1-12 & SWA 1-24 \\
            \hline
             detector & \multicolumn{12}{c|}{\textbf{RetinaNet-R50-1x-0.01-0.0001-36.5}} \\
             \hline
             strategy & \multicolumn{12}{c|}{cyclical lr = (0.01, 0.0001), cycle length = 1 epoch, 12 epochs} \\
             \hline
             bbox AP & 36.2 & 36.4 & 36.7 & 36.7 & 36.8 & 37.0 & 36.9 & 37.0 & 37.2 & 37.2 & 37.2 & 37.2 & 37.2 +0.7 & \textbf{37.8} +1.3 & - \\
             
            \hline
             detector & \multicolumn{12}{c|}{\textbf{RetinaNet-R101-1x-0.01-0.0001-38.5}} \\
             \hline
             strategy & \multicolumn{12}{c|}{cyclical lr = (0.01, 0.0001), cycle length = 1 epoch, 12 epochs} \\
             \hline
             bbox AP & 38.2 & 38.4 & 38.4 & 38.7 & 38.7 & 38.8 & 38.9 & 39.0 & 39.1 & 39.0 & 39.0 & 39.0 & 39.3 +0.8 & \textbf{39.7} +1.2 & - \\
             
             \hline
             detector & \multicolumn{12}{c|}{\textbf{RetinaNet-R101-2x-0.01-0.0001-38.9}} \\
             \hline
             strategy & \multicolumn{12}{c|}{cyclical lr = (0.01, 0.0001), cycle length = 1 epoch, 24 epochs} \\
             \hline
             bbox AP & 39.0 & 39.0 & 38.8 & 39.0 & 39.1 & 39.0 & 38.8 & 39.0 & 38.9 & 39.0 & 38.8 & 38.8 &  &  & \\
             bbox AP & 38.7 & 38.7 & 38.6 & 38.7 & 38.4 & 38.6 & 38.4 & 38.6 & 38.5 & 38.4 & 38.4 & 38.5 & 39.8 +0.9 & \textbf{40.0} +1.1 & \textbf{40.0} +1.1\\

            \hline
        \end{tabular}
    \end{center}
    \vspace{-3mm}
\caption{Performances of SWA RetinaNet.}
\label{table:retinanet}
\end{table*}

\begin{table*}[tpb]
    \begin{center}
        \begin{tabular}{ K{4.0em} | K{1.0em} K{1.0em} K{1.0em} K{1.0em} K{1.0em} K{1.0em} K{1.0em} K{1.0em} K{1.0em} K{1.0em} K{1.0em} K{1.2em} | K{3.8em} K{4.2em} K{4.2em}}
            \hline
             & 1 & 2 & 3 &  4 & 5 & 6 & 7 &  8 & 9 & 10 & 11 & 12 & SWA 1-6 & SWA 1-12 & SWA 1-24 \\
            \hline
             detector & \multicolumn{12}{c|}{\textbf{FCOS-R50-1x-0.01-0.0001-36.6}} \\
             \hline
             strategy & \multicolumn{12}{c|}{cyclical lr = (0.01, 0.0001), cycle length = 1 epoch, 12 epochs} \\
             \hline
             bbox AP & 36.8 & 36.7 & 36.9 & 36.9 & 36.9 & 37.1 & 36.9 & 37.2 & 37.2 & 37.1 & 37.3 & 37.3 & 37.6 +1.0 & \textbf{38.0} +1.4 & - \\
             
            \hline
             detector & \multicolumn{12}{c|}{\textbf{FCOS-R101-1x-0.01-0.0001-39.2}} \\
             \hline
             strategy & \multicolumn{12}{c|}{cyclical lr = (0.01, 0.0001), cycle length = 1 epoch, 12 epochs} \\
             \hline
             bbox AP & 39.2 & 39.1 & 39.5 & 39.3 & 39.3 & 39.4 & 39.5 & 39.5 & 39.4 & 39.4 & 39.4 & 39.4 & 39.9 +0.7 & \textbf{40.3} +1.1 & - \\
             
             \hline
             detector & \multicolumn{12}{c|}{\textbf{FCOS-R101-2x-0.01-0.0001-39.1}} \\
             \hline
             strategy & \multicolumn{12}{c|}{cyclical lr = (0.01, 0.0001), cycle length = 1 epoch, 24 epochs} \\
             \hline
             bbox AP & 39.4 & 39.4 & 39.3 & 39.2 & 39.3 & 39.1 & 39.2 & 39.1 & 39.0 & 39.0 & 39.0 & 38.8 &  &  & \\
             bbox AP & 38.8 & 38.9 & 38.7 & 38.6 & 38.6 & 38.5 & 38.5 & 38.4 & 38.4 & 38.3 & 38.3 & 38.2 & 40.1 +1.0 & \textbf{40.2} +1.1 & 40.0 +0.9\\

            \hline
        \end{tabular}
    \end{center}
    \vspace{-3mm}
\caption{Performances of SWA FCOS.}
\label{table:fcos}
\end{table*}

\begin{table*}[tpb]
    \begin{center}
        \begin{tabular}{ K{4.0em} | K{1.0em} K{1.0em} K{1.0em} K{1.0em} K{1.0em} K{1.0em} K{1.0em} K{1.0em} K{1.0em} K{1.0em} K{1.0em} K{1.2em} | K{3.8em} K{4.2em} K{4.2em}}
            \hline
             & 1 & 2 & 3 &  4 & 5 & 6 & 7 &  8 & 9 & 10 & 11 & 12 & SWA 1-6 & SWA 1-12 & SWA 1-24 \\
            \hline
             detector & \multicolumn{12}{c|}{\textbf{YOLOv3(320)-D53-273e-0.001-0.00001-27.9}} \\
             \hline
             strategy & \multicolumn{12}{c|}{cyclical lr = (0.001, 0.00001), cycle length = 1 epoch, 24 epochs} \\
             \hline
             bbox AP & 27.7 & 27.7 & 27.5 & 27.6 & 27.6 & 27.6 & 27.6 & 27.5 & 27.5 & 27.7 & 27.6 & 27.5 &  &  &  \\
             bbox AP & 27.5 & 27.4 & 27.5 & 27.5 & 27.5 & 27.6 & 27.5 & 27.6 & 27.8 & 27.6 & 27.7 & 27.6 & 28.5 +0.6 & 28.7 +0.8 & \textbf{28.9} +1.0 \\
             
             
            \hline
             detector & \multicolumn{12}{c|}{\textbf{YOLOv3(680)-D53-273e-0.001-0.00001-33.4}} \\
             \hline
             strategy & \multicolumn{12}{c|}{cyclical lr = (0.001, 0.00001), cycle length = 1 epoch, 24 epochs} \\
             \hline
             bbox AP & 33.1 & 33.0 & 33.1 & 33.0 & 33.0 & 33.0 & 32.9 & 32.8 & 32.7 & 32.8 & 32.9 & 32.7 &  &  &  \\
             bbox AP & 32.6 & 32.6 & 32.7 & 32.8 & 33.0 & 32.7 & 32.6 & 32.5 & 32.8 & 32.8 & 32.8 & 32.6 & 34.0 +0.6 & 34.2 +0.8 & \textbf{34.3} +0.9 \\

            \hline
        \end{tabular}
    \end{center}
    \vspace{-3mm}
\caption{Performances of SWA YOLOv3.}
\label{table:yolov3}
\end{table*}

\begin{table*}[tpb]
    \begin{center}
        \begin{tabular}{ K{4.0em} | K{1.0em} K{1.0em} K{1.0em} K{1.0em} K{1.0em} K{1.0em} K{1.0em} K{1.0em} K{1.0em} K{1.0em} K{1.0em} K{1.2em} | K{3.8em} K{4.2em} K{4.2em}}
            \hline
             & 1 & 2 & 3 &  4 & 5 & 6 & 7 &  8 & 9 & 10 & 11 & 12 & SWA 1-6 & SWA 1-12 & SWA 1-24 \\
            \hline
             detector & \multicolumn{12}{c|}{\textbf{VFNet-R50-1x-0.01-0.0001-41.6}} \\
             \hline
             strategy & \multicolumn{12}{c|}{cyclical lr = (0.01, 0.0001), cycle length = 1 epoch, 12 epochs} \\
             \hline
             bbox AP & 41.5 & 41.5 & 41.7 & 41.6 & 41.7 & 41.8 & 41.7 & 41.9 & 41.9 & 41.9 & 41.8 & 41.7 & 42.5 +0.9 & \textbf{42.8} +1.2 & - \\
             
            \hline
             detector & \multicolumn{12}{c|}{\textbf{VFNet-R101-1x-0.01-0.0001-43.0}} \\
             \hline
             strategy & \multicolumn{12}{c|}{cyclical lr = (0.01, 0.0001), cycle length = 1 epoch, 12 epochs} \\
             \hline
             bbox AP & 42.9 & 43.1 & 43.1 & 43.1 & 43.0 & 43.2 & 43.3 & 43.2 & 43.3 & 43.3 & 43.3 & 43.3 & 43.9 +0.9 & \textbf{44.3} +1.3 & - \\
             
             \hline
             detector & \multicolumn{12}{c|}{\textbf{VFNet-R101-2x-0.01-0.0001-43.5}} \\
             \hline
             strategy & \multicolumn{12}{c|}{cyclical lr = (0.01, 0.0001), cycle length = 1 epoch, 24 epochs} \\
             \hline
             bbox AP & 43.7 & 43.4 & 43.4 & 43.4 & 43.4 & 43.5 & 43.3 & 43.4 & 43.3 & 43.4 & 43.1 & 43.2 &  &  & \\
             bbox AP & 43.0 & 43.1 & 42.9 & 43.0 & 42.9 & 42.9 & 42.8 & 42.7 & 42.6 & 42.7 & 42.6 & 42.6 & 44.2 +0.7 & \textbf{44.5} +1.0 & 44.4 +0.9\\
             
             \hline
             detector & \multicolumn{12}{c|}{\textbf{VFNetX(800)-R2(101)-41e-0.01-0.0001-52.2}} \\
             \hline
             strategy & \multicolumn{12}{c|}{cyclical lr = (0.01, 0.0001), cycle length = 1 epoch, 24 epochs} \\
             \hline
             bbox AP & 51.9 & 51.9 & 51.9 & 51.9 & 52.0 & 51.9 & 51.7 & 51.7 & 51.9 & 51.8 & 51.5 & 51.7 &  &  & \\
             bbox AP & 52.0 & 51.8 & 51.5 & 51.4 & 51.4 & 51.6 & 51.4 & 51.6 & 51.4 & 51.4 & 51.5 & 51.4 & 53.0 +0.8 & 53.2 +1.0 & \textbf{53.4} +1.2\\

            \hline
        \end{tabular}
    \end{center}
    \vspace{-3mm}
\caption{Performances of SWA VFNet.}
\label{table:vfnet}
\end{table*}
\vspace{-1mm}

\begin{figure*}[t!]
	\centering
	\begin{subfigure}{0.45\textwidth}
         \centering
         \includegraphics[width=\textwidth]{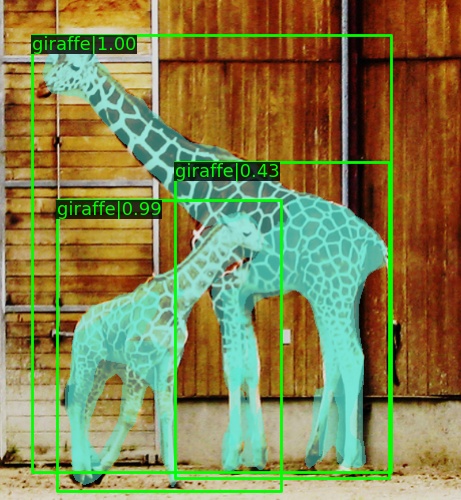}
         \caption{MaskRCNN-R101-2x-0.02-0.0002-40.8-36.6 result}
         \label{fig:maskrcnn_detection}
     \end{subfigure}
     \begin{subfigure}{0.45\textwidth}
         \centering
         \includegraphics[width=\textwidth]{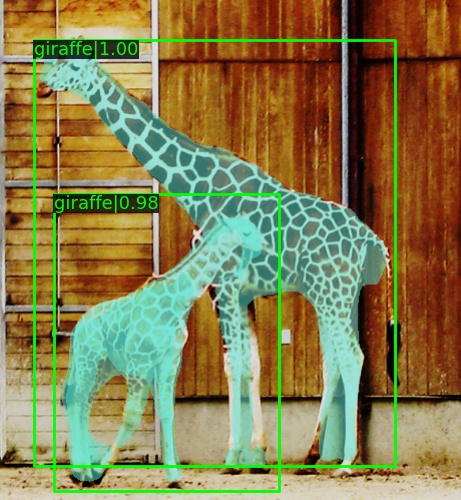}
         \caption{SWA MaskRCNN (41.7 bbox AP and 37.4 segm AP) result}
         \label{fig:swa_maskrcnn_detection}
     \end{subfigure}
     \begin{subfigure}{0.45\textwidth}
         \centering
         \includegraphics[width=\textwidth]{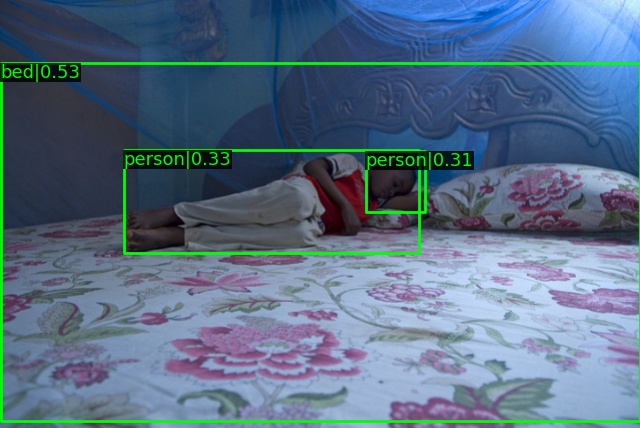}
         \caption{FCOS-R101-2x-0.01-0.0001-39.1 result}
         \label{fig:fcos_detection}
     \end{subfigure}
     \begin{subfigure}{0.45\textwidth}
         \centering
         \includegraphics[width=\textwidth]{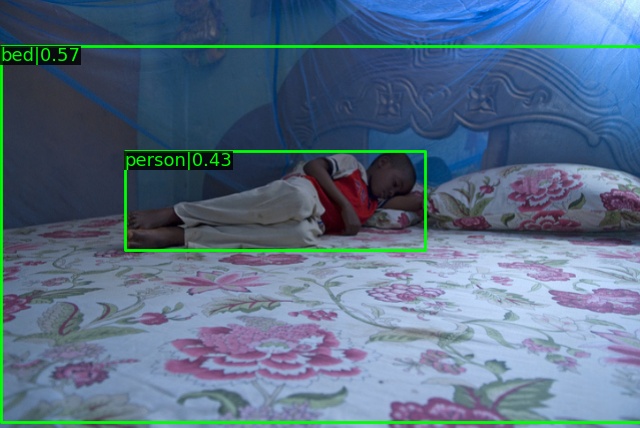}
         \caption{SWA FCOS (40.2 AP) result}
         \label{fig:swa_fcos_detection}
     \end{subfigure}
     \begin{subfigure}{0.45\textwidth}
         \centering
         \includegraphics[width=\textwidth]{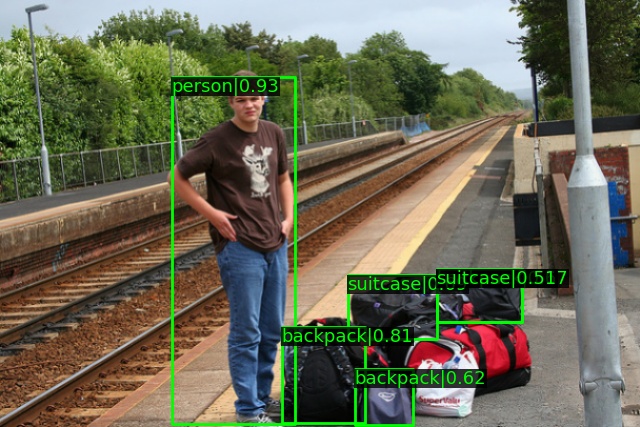}
         \caption{YOLOv3(680)-D53-273e-0.001-0.00001-33.4 result}
         \label{fig:yolov3_detection}
     \end{subfigure}
     \begin{subfigure}{0.45\textwidth}
         \centering
         \includegraphics[width=\textwidth]{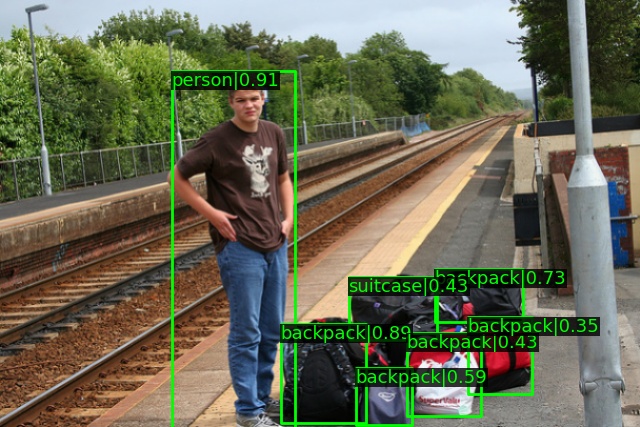}
         \caption{SWA YOLOv3 (34.3 AP) result}
         \label{fig:swa_yolov3_detection}
     \end{subfigure}
\caption{Comparative qualitative examples. Left: detection results of pre-trained models. Right: detection results of corresponding SWA models. The comparison shows that SWA can improve both the object localization and classification accuracy, resulting in less false positives and higher recall rates.}
\label{fig:qualitative_examples}
\end{figure*}

\begin{figure*}[t!]
	\centering
	\begin{subfigure}[a]{0.48\textwidth}
         \centering
         \includegraphics[width=\textwidth]{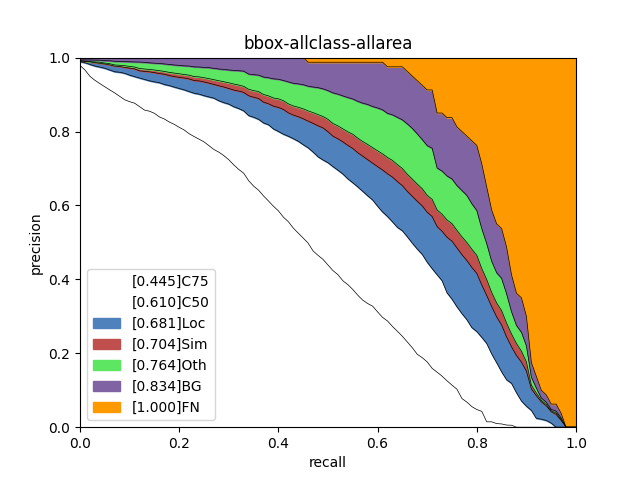}
         \caption{Pre-trained Mask RCNN Bbox Results (40.8 AP) Analysis}
         \label{fig:maskrcnn_analysis_a}
     \end{subfigure}
     \begin{subfigure}[a]{0.48\textwidth}
         \centering
         \includegraphics[width=\textwidth]{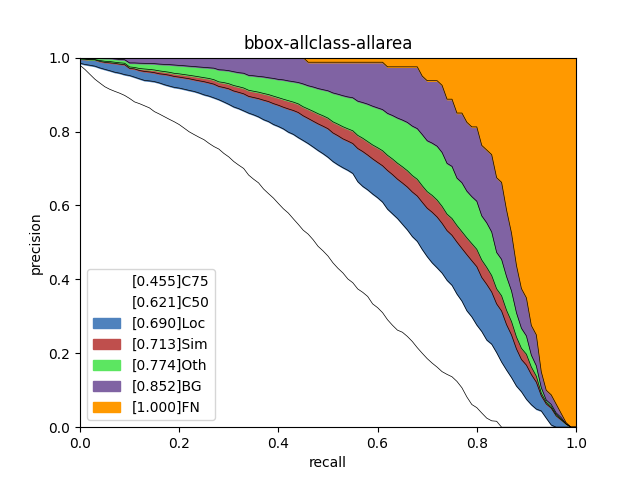}
         \caption{SWA Mask RCNN Bbox Results (41.7 AP) Analysis}
         \label{fig:maskrcnn_analysis_c}
     \end{subfigure}
     \begin{subfigure}[a]{0.48\textwidth}
         \centering
         \includegraphics[width=\textwidth]{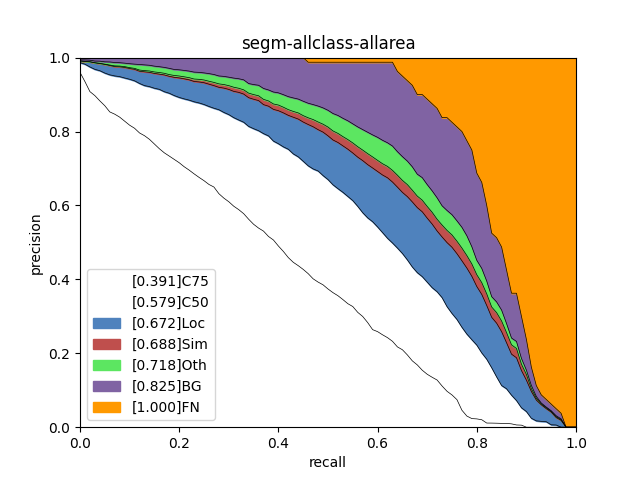}
         \caption{Pre-trained Mask RCNN Segm Results (36.6 AP) Analysis}
         \label{fig:maskrcnn_analysis_b}
     \end{subfigure}
     \begin{subfigure}[a]{0.48\textwidth}
         \centering
         \includegraphics[width=\textwidth]{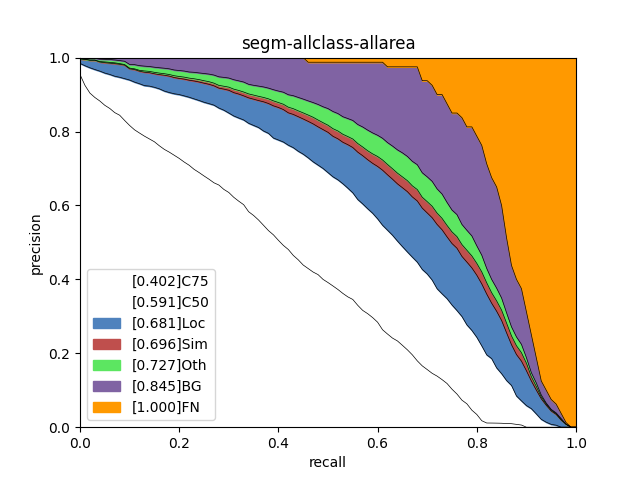}
         \caption{SWA Mask RCNN Segm Results (37.4 AP) Analysis}
         \label{fig:maskrcnn_analysis_d}
     \end{subfigure}
\caption{Analysis of MaskRCNN-R101-2x-0.02-0.0002-40.8-36.6 and its SWA model.}
\label{fig:maskrcnn_analysis}\vspace{-3mm}
\end{figure*}

\vspace{2mm}
\begin{figure*}[t!]
	\centering
     \begin{subfigure}[a]{0.48\textwidth}
         \centering
         \includegraphics[width=\textwidth]{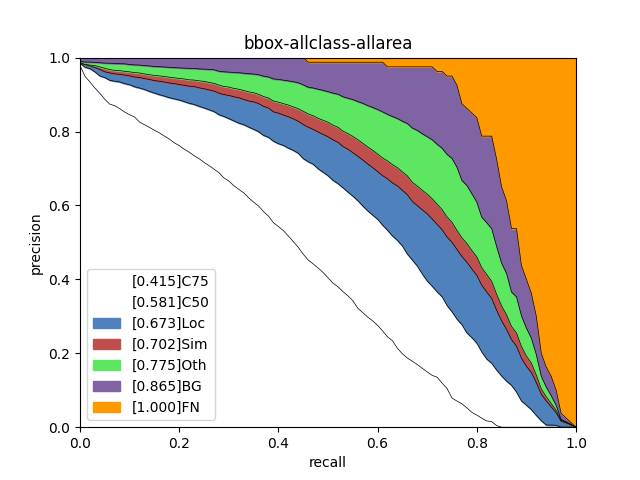}
         \caption{Pre-trained FCOS Results (39.1 AP) Analysis}
         \label{fig:fcos_analysis_a}
     \end{subfigure}
     \begin{subfigure}[a]{0.48\textwidth}
         \centering
         \includegraphics[width=\textwidth]{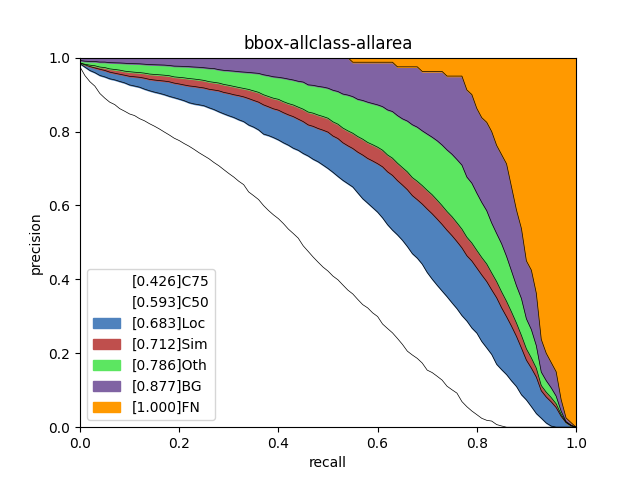}
         \caption{SWA FCOS Results (40.2 AP) Analysis}
         \label{fig:fcos_analysis_b}
     \end{subfigure}
\caption{Analysis of FCOS-R101-2x-0.01-0.0001-39.1 and its SWA model.}
\label{fig:fcos_analysis}\vspace{-3mm}
\end{figure*}

\subsection{Ablation Study}
We select Mask RCNN~\cite{maskRCNN} as our study detector to search for the proper way of using SWA in object detection as well as instance segmentation. 

We first download the pre-trained model, MaskRCNN-R101-2x-0.02-0.0002-40.8-36.6, and its configuration file from the MMDetection model zoo as our starting point.

Then we train the model for another 24 or 48 epochs with different learning rate strategies. The first kind is a fixed learning rate schedule where 0.02, 0.002 and 0.0002 are selected. Note that these learning rates correspond to those ones used in different stages of the training of the pre-trained model. The second strategy is a cyclical learning rate schedule. As shown in Figure~\ref{fig:cyclical_lr}, in each cycle the learning rate starts with a large value $lr_{max}$ and then relatively rapidly decreases to a minimum value $lr_{min}$ before jumping to the maximum value again. Note that the learning rate decreases at each iteration rather than at each epoch. In this study, we adopt the cosine annealing learning rate schedule, choose two sets of ($lr_{max}$, $lr_{min}$), \ie (0.01, 0.0001) and (0.02, 0.0002), and opt for 1 epoch as the cycle length. 

Finally, we average different numbers (6, 12, 24 and 48) of the new checkpoints as our final SWA models and evaluate their performances on the COCO val2017. Note that since batch normalization layers in backbones are frozen~\cite{mmdetection}, we do not need to follow the original SWA paper to run an additional pass over the data to compute the new statistics.

The results are presented in Table~\ref{table:maskrcnn}. As mentioned above, we have tried five different training strategies and they are divided into two groups. For the fixed learning rate group, we can see that learning rates have a great impact on the performance of each SGD epoch. Specifically, with the learning rate being 0.02, each SGD epoch performs much worse than the pre-trained model, \eg 33.0 $\sim$ 34.0 bbox AP vs 40.8 bbox AP. By contrast, when the learning rate is 0.0002, each SGD epoch performs comparably to the pre-trained model. 

Although the performances achieved by each SGD epoch with different learning rates vary significantly, surprisingly, the SWA models that are obtained by averaging the certain number of checkpoints under each training strategy achieve quite similar results. We can see that in the SWA 1-12 column of Table~\ref{table:maskrcnn}, all those three SWA models attain about 40.5 bbox AP and 36.5 mask AP. However, these results are inferior to that of the starting model, indicating that the constant learning rate strategy does not work well.

In comparison, the cyclical learning rate group achieves more stable results in each SGD epoch and their SWA models reach much better results. It can be seen that the learning rate range (0.02, 0.0002) performs better than the range (0.01, 0.0001), showing that the learning rates used in the pre-training phase already works well. Taking a closer look at the results of the (0.02, 0.0002) range, its SWA 1-12 model achieves 41.7 bbox AP and 37.4 mask AP, improving 0.9 bbox AP and 0.8 mask AP over the pre-trained model respectively.  Moreover, SWA 1-12 model performs better than SWA 1-6 model, and is comparable to both SWA 1-24 model and SWA 1-48 model. This indicates that training another 12 epochs is enough for generating a good SWA model, especially when considering the trade-off between computation overhead and gain.

Comparing those results in Table~\ref{table:maskrcnn}, we can infer one workable strategy of applying SWA in training better object detectors. That is, \textbf{after the conventional training of an object detector with the initial learning rate $lr_{ini}$ and the ending learning rate $lr_{end}$, train it for an extra 12 epochs using the cyclical learning rates ($lr_{ini}$, $lr_{end}$) for each epoch, and then average these 12 checkpoints as the final detection model}.

Based on the observations above, we also tried training SWA Mask RCNN from scratch (the backbone is pre-trained on ImageNet~\cite{ImageNet}). We first train a raw Mask RCNN model for 16 epochs with the learning rate 0.02, getting the model MaskRCNN-R101-16e-0.02-0.02-33.4-30.8. Then, we train it for another 12 epochs with cyclical learning rates (0.02, 0.0002). Finally, we average these 6 or 12 checkpoints as the SWA models. As shown in the last part of Table~\ref{table:maskrcnn}, we can see that this SWA 1-12 model achieves 41.7 bbox AP and 37.4 mask AP, which is the same as the SWA 1-12 model obtained by training the MaskRCNN-R101-2x-0.02-0.0002-40.8-36.6. This shows that such a mixed training strategy can also generate a better object detector and can be used to train a new object detector from scratch.

\subsection{Main Results}
To verify the effectiveness of the strategy that we discovered for performing SWA in object detection, we apply it to various object detectors with different backbones, including Mask RCNN, Faster RCNN, RetinaNet, FCOS, YOLOv3 and our VFNet. Results are presented in Table~\ref{table:maskrcnn_res}, Table~\ref{table:fasterrcnn}, Table~\ref{table:retinanet}, Table~\ref{table:fcos}, Table~\ref{table:yolov3} and Table~\ref{table:vfnet}, respectively. From these results, we can see that SWA with our training police consistently improves the performances of these detectors by $\sim$1.0 AP, irrespective of whether their original performance is high or low. This is very encouraging and makes us excited to share the discovery to the community. 

Comparative qualitative examples can be viewed in Figure~\ref{fig:qualitative_examples}. Comparing these detection examples, we can see that SWA improves both the object localization and object classification accuracy, resulting in less false positives and higher recall rates.

\subsection{Analysis}
To further understand where the improvements that SWA brings come from, we analyze the results of Mask RCNN and FCOS. Following the practice in the paper, \textit{Diagnosing Error in Object Detectors}~\cite{diagnose}, we plot the breakdown of errors of the pre-trained Mask RCNN (MaskRCNN-R101-2x-0.02-0.0002-40.8-36.6) and its SWA model, as well as FCOS (FCOS-R101-2x-0.01-0.0001-39.1) and its SWA model. The plots that are generated by COCO API are separately shown in Figure~\ref{fig:maskrcnn_analysis} and Figure~\ref{fig:fcos_analysis}. The detailed explanation of such plots can be found on the COCO dataset webpage~\cite{COCO_analysis}. In short, each plot is a series of precision recall (PR) curves where each PR curve is guaranteed to be strictly higher than the previous as the evaluation setting becomes more permissive, and the area under each curve corresponds to the AP (shown in brackets in the legend).

Comparing Figure~\ref{fig:maskrcnn_analysis_a} and Figure~\ref{fig:maskrcnn_analysis_c}, Figure~\ref{fig:maskrcnn_analysis_b} and Figure~\ref{fig:maskrcnn_analysis_d}, as well as Figure~\ref{fig:fcos_analysis_a} and Figure~\ref{fig:fcos_analysis_b}, we can infer that SWA improves not only the object localization accuracy but also the object classification accuracy. For example, Figure~\ref{fig:maskrcnn_analysis_a} shows the pre-trained Mask RCNN achieves overall AP at IoU=0.75 is 44.5, but SWA Mask RCNN improves this number by 1.0 to 45.5 AP, indicating SWA improves the localization accuracy. Similarly, when the localization errors are ignored, which is represented by Loc in the legend, the pre-trained Mask RCNN achieves 68.1 AP, but SWA Mask RCNN reachs 69.0 AP, which implies the object classification accuracy is also enhanced by SWA. The similar comparative results can also be seen for FCOS.

\section{Conclusion}
In this report, we systematically investigate the effects of applying SWA to object detection and instance segmentation. We find that training an object detector for another 12 epochs with cyclical learning rates and averaging these 12 checkpoints as the final model can improve $\sim$1.0 AP for this detector on the challenging COCO benchmark. Our extensive experiments show that this technique works well with various object detectors, including Mask RCNN, Faster RCNN, RetinaNet, FCOS, YOLOv3 and VFNet. We hope our work can make more researchers know this simple but effective recipe and help them train better object detectors. 

{\small
\bibliographystyle{unsrt}
\bibliography{swa}
}

\end{document}